\documentclass[11pt]{article}

\usepackage{acl}

\usepackage{times}
\usepackage{latexsym}
\usepackage{bbding} 
\usepackage[T1]{fontenc}

\usepackage[utf8]{inputenc}
\usepackage{subcaption}
\usepackage[ruled,linesnumbered]{algorithm2e}
\usepackage{microtype}

\usepackage{inconsolata}

\usepackage{amsmath, amssymb}
\usepackage{graphicx}
\usepackage{booktabs}
\usepackage{multirow}
\usepackage{makecell}
\usepackage{enumitem}
\setlist[itemize]{nosep}
\setlist[enumerate]{nosep}

\usepackage{caption}
\usepackage[most]{tcolorbox} 
\tcbuselibrary{minted, skins} 
\usepackage{float}

\newtcolorbox{promptbox}[1][]{
  enhanced,
  boxrule=0.8pt,
  colback=gray!5,
  colframe=gray!70,
  arc=3mm,
  auto outer arc,
  breakable=false, 
  before upper={%
    \setlist[itemize]{nosep}%
    \def\sep{\vspace{5pt}}%
  },
  #1
}

\title{DeepSynth-Eval: Objectively Evaluating Information Consolidation in Deep Survey Writing}

\author{
  \textbf{Hongzhi Zhang\textsuperscript{\dag}\thanks{\ \ Equal contribution.}}, 
  \textbf{Yuanze Hu\textsuperscript{\ddag}\footnotemark[1]},
  \textbf{Tinghai Zhang\textsuperscript{\dag}\footnotemark[1]},
  \textbf{Jia Fu\textsuperscript{\dag}\footnotemark[1]},
  \textbf{Tao Wang\textsuperscript{\dag}\footnotemark[1]},
  \\
  \textbf{Junwei Jing\textsuperscript{\dag}},
  \textbf{Zhaoxin Fan\textsuperscript{\ddag}},
  \textbf{Qi Wang\textsuperscript{\dag}},
  \textbf{Ruiming Tang\textsuperscript{\dag\,\Envelope}}, 
  \textbf{Han Li\textsuperscript{\dag}},
  \textbf{Guorui Zhou\textsuperscript{\dag}},
  \textbf{Kun Gai\textsuperscript{\dag}}
  \\
  \textsuperscript{\dag}Kuaishou Technology\\
  \textsuperscript{\ddag}Beijing Advanced Innovation Center for Future Blockchain and Privacy Computing
}
\allowdisplaybreaks[4]

\begin{document}
\setlength{\abovedisplayskip}{5pt}
\setlength{\belowdisplayskip}{5pt}
\maketitle
\begingroup
  \renewcommand\thefootnote{\Envelope} 
  \footnotetext{\space \space Corresponding author: \href{mailto:tangruiming@kuaishou.com}{tangruiming@kuaishou.com}}
\endgroup
\begin{abstract}
The evolution of Large Language Models (LLMs) towards autonomous agents has catalyzed progress in Deep Research. While retrieval capabilities are well-benchmarked, the post-retrieval synthesis stage—where agents must digest massive amounts of context and consolidate fragmented evidence into coherent, long-form reports—remains under-evaluated due to the subjectivity of open-ended writing. 

To bridge this gap, we introduce DeepSynth-Eval, a benchmark designed to objectively evaluate information consolidation capabilities. We leverage high-quality survey papers as gold standards, reverse-engineering research requests and constructing "Oracle Contexts" from their bibliographies to isolate synthesis from retrieval noise. We propose a fine-grained evaluation protocol using General Checklists (for factual coverage) and Constraint Checklists (for structural organization), transforming subjective judgment into verifiable metrics. Experiments across 96 tasks reveal that synthesizing information from hundreds of references remains a significant challenge. Our results demonstrate that agentic ``plan-then-write'' workflows significantly outperform single-turn generation, effectively reducing hallucinations and improving adherence to complex structural constraints. 
\end{abstract}

\section{Introduction}
The evolution of Large Language Models (LLMs) \cite{deepseekai2025deepseekr1incentivizingreasoningcapability} marks a paradigm shift from fundamental token completion to sophisticated task completion via tool augmentation\cite{yao2023react}. This transition has notably accelerated progress in Deep Research\cite{OpenAIdeepresearch,Geminideepresearch,shao-etal-2024-assisting}, where agents are tasked with not only retrieving information but also synthesizing it into comprehensive, survey-style reports.
While Information Retrieval (IR) has robust benchmarks such as GAIA\cite{mialon2023gaiabenchmarkgeneralai}, HLE\cite{phan2025humanitysexam} and others\cite{wei2025browsecompsimplechallengingbenchmark,wu-etal-2025-webwalker}, the latter stage—Post-Retrieval Synthesis—remains under-evaluated. In this stage, models must digest massive amounts of retrieved contexts, consolidate fragmented evidence, and organize them into a coherent narrative. Evaluating this capability is challenging. Unlike short-form QA tasks like GAIA and HLE where answers can be checked against ground-truth facts, deep survey reports are long, lack a unique reference, and vary widely in content selection, structure, and writing style. Consequently, current practices rely heavily on subjective human judgments or LLM-as-a-judge scores\cite{LLM-as-a-judge}, which are costly, unstable, and prone to reward hacking\cite{wang-etal-2024-large-language-models-fair}. Furthermore, end-to-end evaluations often conflate retrieval noise with synthesis failures, obscuring the model's true ability to consolidate information.

To objectively evaluate this information consolidation capability, we propose to assess deep synthesis performance via survey papers. High-quality survey papers can be viewed as expert-written research reports that serve as ``gold standards'' of synthesis. For each survey, we first ``reverse-engineer'' a corresponding research request and construct the input context from its bibliography. We then manually extract a set of checklists that specify what a good report must cover under this request, including representative methods, key design choices, important datasets, and other domain-specific points. These checklists transform open-ended report evaluation into verifying whether the model's output accurately covers the required items, enabling a more objective measurement of completeness and factual correctness.

\begin{figure*}[t]
    \centering
    \vspace{-1.5cm}
    \includegraphics[width=0.95\linewidth]{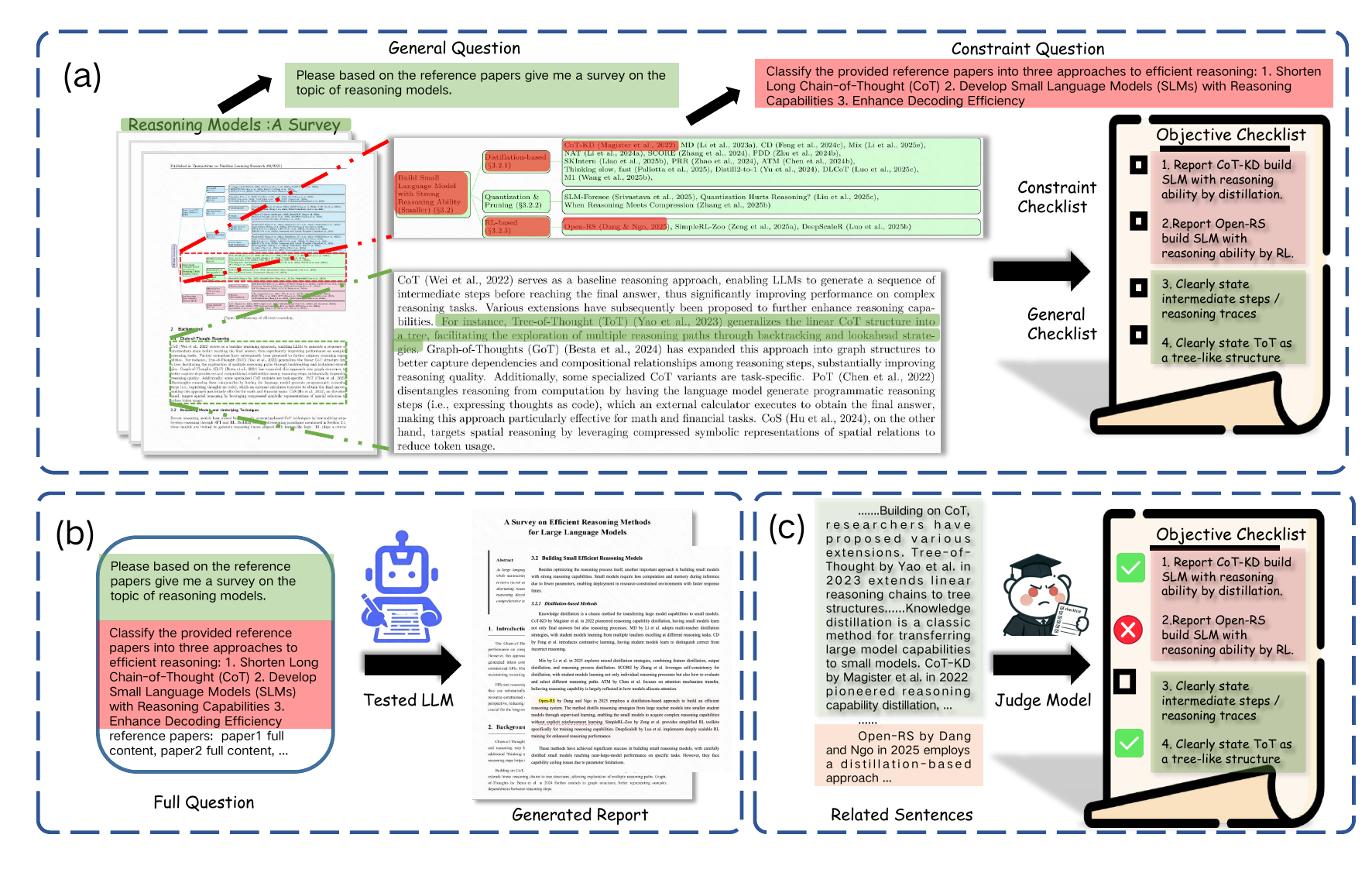}
    \vspace{-0.5cm}
    \caption{An illustration of DeepSynth-Eval. (a) From a reference survey, we derive a general prompt and constraint questions, and construct corresponding general/constraint checklists for evaluation. (b) The full question (prompt + constraints + reference papers) is fed to a tested LLM to generate a report. (c) A judge model verifies each checklist item on the generated report, converting subjective synthesis evaluation into objective item-wise verification.}

    \vspace{-0.50cm}
    
    \label{fig:workflow_comparison}
\end{figure*}

Beyond factual coverage, a hallmark of deep synthesis is structural organization—for example, consolidating methods into comparative tables or grouping them into meaningful taxonomies. However, such organization is inherently subjective and flexible: there are diverse valid ways to categorize works or select attributes for comparison. Consequently, expecting a model to spontaneously replicate the exact structure of a reference survey is unreasonable. To resolve this ambiguity and make organizational abilities objectively comparable, we extract a canonical set of structural constraints from each survey (e.g., specific taxonomies to adopt or attributes to include in a table). By explicitly imposing these constraints in the synthesis prompt, we reduce editorial freedom, which allows us to compare the organization of content in corresponding sections. We refer to these requirements as Constraint Questions and the corresponding checklists as Constraint Checklists.

Building on these ideas, we construct \textbf{DeepSynth-Eval (DSE)}, a benchmark designed for the objective evaluation of \textbf{deep synthesis} capabilities. DSE comprises 96 queries derived from high-quality surveys across diverse domains. For each query, the benchmark provides the reconstructed research request, a comprehensive \textbf{Oracle Context} (constructed from the bibliography), and a set of fine-grained general and constraint checklists. Given a model-generated report, our pipeline automatically aligns the content with checklist items to compute coverage, precision, and structural constraint satisfaction scores. Compared to directly judging unstructured free-form text, this key-point verification reduces the knowledge burden on judge models, supports fine-grained and discriminative scoring, and makes the evaluation more robust and reproducible—providing a reliable foundation for training and improving deep synthesis systems.

Our main contributions are: \begin{itemize} 

\item \textbf{Objective Checklist Metric.} We address the challenge of evaluating long-form generation by introducing a checklist-based metric derived from gold-standard surveys. This method decomposes synthesis quality into verifiable factual coverage and structural constraint adherence, providing a robust standard for measuring deep research capabilities.
\item \textbf{DeepSynth-Eval (DSE) Benchmark.} We introduce DSE, a benchmark comprising 96 tasks equipped with Oracle Contexts. This design isolates post-retrieval synthesis from retrieval noise, ensuring reproducibility and preventing ground-truth leakage.
\item \textbf{Evaluation and Open Source.} We establish a rigorous baseline by benchmarking end-to-end and agentic workflows. We find that while agentic methods dominate, synthesizing context from over 100 references remains a formidable bottleneck. We open-source the entire framework to accelerate research in this direction.\footnote{\url{https://github.com/Kwai-Klear/DeepSynth-Eval}}
\end{itemize}

\section{Related Work}

The paradigm of LLMs is shifting from closed-ended tasks---exemplified by benchmarks like MMLU~\cite{hendrycks2021measuringmassivemultitasklanguage}, GPQA~\cite{rein2023gpqagraduatelevelgoogleproofqa}, and Humanity's Last Exam~\cite{wang2024mmluprorobustchallengingmultitask}---to complex \textbf{Deep Research}. Unlike these static evaluations, Deep Research requires integrating reading, searching, and writing. Existing works fall into three categories.

\paragraph{Long-Context Understanding}
Serving as the cognitive foundation for deep research, Long-Context Understanding has garnered significant attention; however, current evaluations are often confined to information extraction rather than logical restructuring. 
Existing benchmarks, such as LongBench~\cite{bai2024longbenchbilingualmultitaskbenchmark}, InfiniteBench~\cite{zhang-etal-2024-bench}, and L-Eval~\cite{an2023levalinstitutingstandardizedevaluation}, primarily assess the model's capacity for information retention within extended context windows. 
However, the task formulations in these benchmarks are largely restricted to ``Needle-in-a-Haystack'' scenarios or local extractive QA. 
In such settings, models can typically succeed by locating and reciting specific segments within the context, without the necessity for global association or cross-document reasoning to synthesize unstructured evidence dispersed across numerous documents.

\paragraph{Agentic Retrieval and Fact-Finding}
Transitioning from passive information processing to active information acquisition, Deep Search benchmarks (e.g., GAIA~\cite{mialon2023gaiabenchmarkgeneralai}, BrowseComp~\cite{wei2025browsecompsimplechallengingbenchmark}) focus on evaluating an agent's ability to utilize tools for solving complex problems. 
However, these evaluations typically involve \textit{low information throughput}, where retrieving a few key pages suffices. 
Consequently, they fall short of measuring the comprehensive ability to screen and synthesize massive information into coherent long-form reports.
To address scale, WideSearch~\cite{wong2025widesearchbenchmarkingagenticbroad} targets breadth-oriented acquisition. 
Yet, despite the increased throughput, the task remains limited to \textbf{Structured Information Extraction} (e.g., compiling tables). 
This primarily tests recall in repetitive tasks, lacking the depth required for cross-document comparative analysis and the integration of heterogeneous information needed in deep research.

\paragraph{End-to-End Research System Evaluations}

Deep Research tasks inherently require agents to complete the full workflow from open-web searching to long-form report generation (Search-Read-Write). 
To align with this real-world application scenario, recent works have adopted an end-to-end evaluation paradigm; however, this setting introduces confounding variables that are difficult to isolate. For instance, PDR-Bench~\cite{liang2025personalizeddeepresearchbenchmarks} and DeepResearchBench~\cite{du2025deepresearchbenchcomprehensivebenchmark} focus on personalization alignment and LLM-based subjective scoring for complex tasks, respectively, whereas ReportBench~\cite{li2025reportbenchevaluatingdeepresearch} assesses the Search-Read-Write capabilities of end-to-end systems by reverse-engineering arXiv survey papers. 

Despite these significant explorations in system-level assessment, this open-web evaluation methodology faces two major methodological challenges.
The first challenge is the \textbf{Coupling of Retrieval and Generation}. 
The \textit{dynamic drift of rankings} causes context variations over time, precipitating a \textit{reproducibility crisis} where it is indistinguishable whether gains stem from model enhancements or better retrieval luck. 
Furthermore, retrieval noise (e.g., ads, broken links) prevents accurate measurement of the model's \textit{intrinsic cognitive capability} in processing complex information.
The second is the \textbf{Subjectivity of Evaluation}. 
Existing methods predominantly rely on LLMs to assign holistic scores (score-based) to reports, lacking objective verification of specific factual point coverage and remaining susceptible to biases inherent in the judge models.

\section{Benchmark Construction}
\label{section:benchmark_construction}

\subsection{Design Philosophy}
We propose to construct a benchmark utilizing high-quality survey papers as reference answers. Formally, given a written survey report $R$, we first reverse-engineer a query $q_g$ for which the report $R$ serves as the answer; we term this the \textbf{General Question} ($q_g$) (e.g., ``Summarize recent advances in reasoning models'').

Intuitively, one might evaluate the model's performance by directly comparing the consistency between the generated report $\hat{R} = \mathcal{M}(q_g)$ and the original survey $R$. However, in practice, directly comparing two lengthy, unstructured reports is computationally and metrically challenging.
To address this, we further extract a set of \textbf{checklists} from $R$, where each item corresponds to a specific, verifiable fact. These checklists specify the essential content that a comprehensive report should cover under the given question $q_g$, thereby transforming the evaluation into a checklist verification task.

Survey writing is inherently subjective; different authors may adopt valid but distinct strategies to organize the same body of literature. For instance, benchmarks for Large Language Models (LLMs) could be categorized by the \textit{capability domain} (e.g., reading comprehension, math, coding) or by the \textit{task format} (e.g., multiple-choice questions vs. open-ended generation). Consequently, without constraints, the structural organization of a generated report is often incomparable to the reference.

To address this, we extract specific structural requirements from the reference survey and append them as \textbf{Constraint Questions} ($q_c$) to the general prompt. For example, we might explicitly require the model to classify efficient reasoning methods into three specific categories: \textit{(1) Shortening Chain-of-Thought, (2) Developing Small LLMs with reasoning capabilities, and (3) Enhancing decoding efficiency.} In this manner, we can objectively track whether the system correctly organizes representative works into the corresponding classes.

We term the extracted structural requirements as \textbf{Constraint Questions} ($q_c$) and the corresponding evaluation points as \textbf{Constraint Checklists}. Note that these checklists are valid for evaluation only when $q_c$ is provided. By imposing $q_c$, we effectively reduce the editorial freedom of the synthesis task, rendering the results comparable. Furthermore, this design allows for a more rigorous assessment of the model's {instruction-following} capabilities within complex writing tasks.

Overall, given the composite input comprising the general question $q_g$ and a set of constraint questions $q_c$, the model synthesizes a report by processing the reference bibliography of the original survey. Finally, an evaluation system verifies the generated report $\hat{R}$ against the comprehensive set of checklists $\mathcal{C} = \mathcal{C}_{gen} \cup \mathcal{C}_{con}$ to assess the satisfaction of each requirement.

\begin{figure}
    \centering
    \includegraphics[width=\columnwidth]{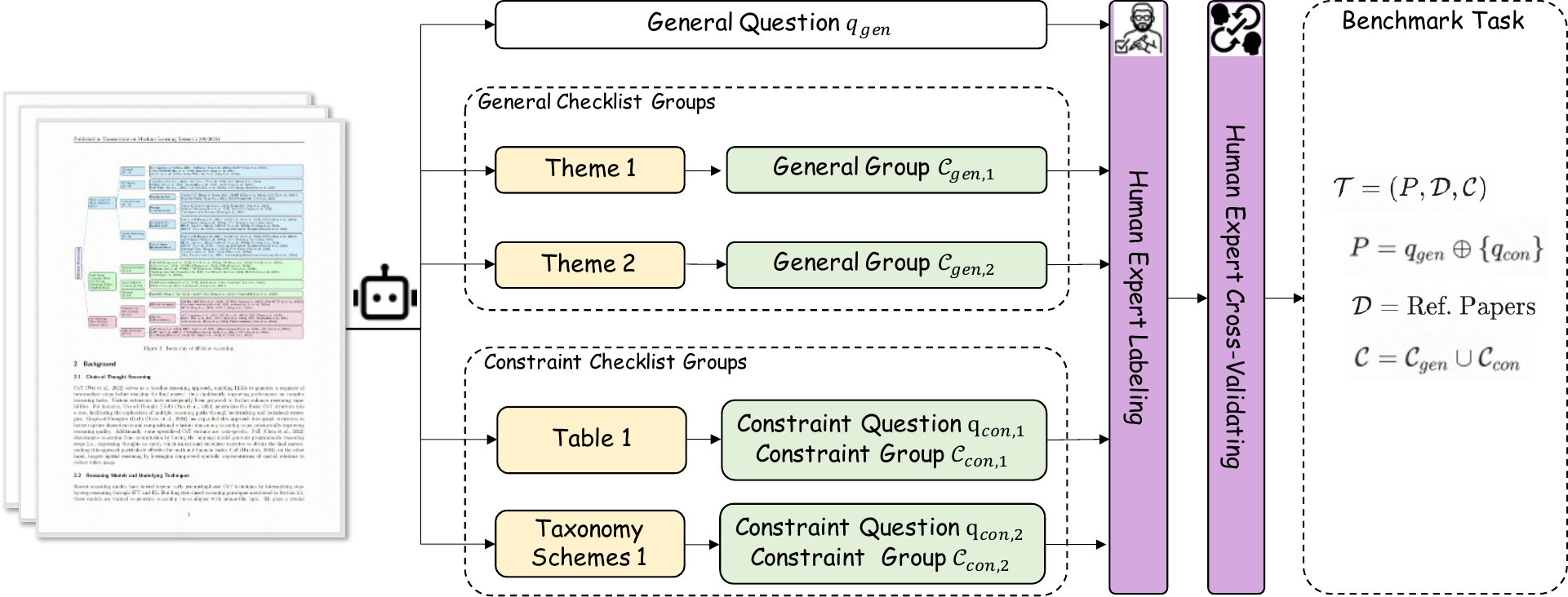}
    \vspace{-0.3cm}

    \caption{Overview of the DeepSynth-Eval construction process. Candidates generated by LLMs undergo human verification and editing to ensure quality.} 
    \label{figure:data_synthesis_workflow}
    \vspace{-0.5cm}
\end{figure}

\subsection{Automated Construction Pipeline}
The construction of this writing benchmark itself could be very labor consuming. In this paper, we utilize the LLM to generate candidates and with human labor to review and cross validate the dataset. 

To construct the benchmark efficiently while maintaining high quality, we employ a multi-stage generation pipeline utilizing LLMs.

\noindent \textbf{General Question Generation ($q_{gen}$).} 
This step is straightforward. We treat the survey title and abstract as the core intent and employ an LLM to directly reverse-engineer the corresponding research request. This produces the general question $q_{gen}$ (e.g., ``Provide a comprehensive survey on ...''), which serves as the foundation of the task.

\noindent \textbf{General Checklist Generation ($\mathcal{C}_{gen}$).} 
Directly generating a comprehensive set of fine-grained checklists from a lengthy document is challenging for current models, often resulting in low recall or hallucinations. To mitigate this, we adopt a hierarchical \textbf{Topic Anchor} strategy. 
First, we prompt the LLM to synthesize several distinct \textit{topics} based on the survey's section headers and content (e.g., ``Core Algorithms'', ``Evaluation Metrics'', ``Future Directions''). 
Then, using these topics as anchors, we invoke the LLM to generate specific checklist items for each topic individually. This divide-and-conquer approach yields $K$ groups of general checklists, denoted as $\mathcal{C}_{gen} = \bigcup_{k=1}^{K} \mathcal{C}_{gen, k}$.

\noindent \textbf{Constraint Pair Generation ($q_{con}, \mathcal{C}_{con}$).} 
For structural constraints, we employ a \textbf{Structure Anchor} approach to generate coupled pairs of constraint questions and checklists.
\begin{enumerate}
    \item \textbf{Anchor Extraction:} The model first scans the survey to identify key organizational elements, such as specific taxonomies (classification schemes) or summary tables.
    \item \textbf{Guidance Generation ($q_c$):} Based on an extracted anchor (e.g., a table summarizing reasoning datasets), the model generates a \textbf{Constraint Question} $q_c$ that acts as explicit guidance. For instance, if the anchor is a dataset table, $q_c$ might be: \textit{``List a table summarizing recent reasoning datasets, specifying their data scale, question source, and answer source.''}
    \item \textbf{Checklist Extraction:} Simultaneously, the model extracts the corresponding cell values or classification criteria from the anchor to form a group of \textbf{Constraint Checklists} $\mathcal{C}_{con}$.
    This process yields $M$ groups of constraint checklists, denoted as $\mathcal{C}_{con} = \bigcup_{m=1}^{M} \mathcal{C}_{con, m}$.
\end{enumerate}

\subsection{Human Verification}

In the construction of DeepSynth-Eval, the automated workflow initially yielded 322 candidate tasks. To ensure benchmark reliability, we implemented a rigorous two-stage human verification process. The annotation team consisted of six experts specializing in computer science.
Through the first round of manual screening and editing, we retained 129 instances. Subsequently, following a second round of expert cross-validation and final refinement, we curated a final set of 96 high-quality instances to constitute the benchmark.

During the screening, we identified and discarded problematic instances characterized by common generation issues, such as: (1) Over-specification, where constraint questions were so detailed that the corresponding checklists became trivially inferable; and (2) Contextual Ambiguity, particularly in table-based extractions, where visual symbols (e.g., $\checkmark$ or $\times$) were extracted without sufficient context to permit verification.

\subsection{Task Formalization}

Formally, we define a deep synthesis evaluation task in DeepSynth-Eval as a tuple $\mathcal{T} = (P, \mathcal{D}, \mathcal{C})$, where:

\begin{itemize}
    \item $P$ denotes the \textbf{Unified Input Prompt}. It is constructed by concatenating the general question $q_{gen}$ with a set of $M$ constraint questions $\{q_{con}^{(1)}, \dots, q_{con}^{(M)}\}$. Formally:
    $$ P = q_{gen} \oplus q_{con}^{(1)} \oplus \dots \oplus q_{con}^{(M)} $$
    where $\oplus$ represents the string concatenation operation.
    
    \item $\mathcal{D}$ represents the \textbf{Oracle Context}, containing the full text of the reference bibliography associated with the original survey. This serves as the closed-world knowledge boundary for the task.
    
    \item $\mathcal{C}$ is the \textbf{Evaluation Standard}, composed of the union of general checklist groups and constraint checklist groups: $\mathcal{C} = \mathcal{C}_{gen} \cup \mathcal{C}_{con}$.
\end{itemize}

\subsection{Evaluation Metric}
During inference, the model $\mathcal{M}$ receives the unified prompt $P$ and the context $\mathcal{D}$ to synthesize a report $\hat{R} = \mathcal{M}(P, \mathcal{D})$. 
The quality of $\hat{R}$ is then quantified by assessing its satisfaction of the checklist set $\mathcal{C}$. Specifically, for each checklist item $c \in \mathcal{C}$, we employ a judge model to verify its presence in $\hat{R}$, computing a final score based on the satisfaction rate.

Specifically, the judge model evaluates each requirement and assigns a reward $r$ based on three distinct verification states:
\begin{equation*}
\resizebox{\columnwidth}{!}{
$\displaystyle r=\begin{cases}
1, & \text{if mentioned correctly (Correctness)};\\
0, & \text{if not mentioned (Omission)};\\
-1, & \text{if mentioned incorrectly (Hallucination)}.
\end{cases}$}
\end{equation*}

\noindent \textbf{Fault-Tolerant Scoring.}
Given the open-ended nature of survey writing, exhaustive enumeration of every possible detail is often unnecessary; capturing the majority of representative works usually suffices for a high-quality report. To reflect this, we introduce a \textbf{saturation threshold} $\theta_k$ for each checklist group $k$.

Let $N_k$ be the total number of items in group $k$, and $S_k$ be the cumulative reward sum obtained by the generated report (aggregating rewards of $1, 0$ and $-1$). The normalized score for group $k$ is calculated as:
\begin{equation}
    \text{Score}_k = \min\!\left( 1, \frac{S_k}{\theta_k} \right)
\end{equation}
where $\theta_k \leq N_k$. This mechanism implies that once the model satisfies a sufficient number of requirements (i.e., $S_k \geq \theta_k$), it receives a full score for that group. For instance, in a checklist group listing $N_k=10$ representative methods, we may set the threshold at $\theta_k=8$, meaning that correctly covering 8 out of 10 key methods is considered ``good enough'' to achieve the maximum score.

\paragraph{Aggregated Metrics.}
To provide a fine-grained analysis of model performance, we report metrics across three dimensions: \textbf{General Score} ($\mathcal{S}_{gen}$), which measures the coverage of core research content; \textbf{Constraint Score} ($\mathcal{S}_{con}$), which evaluates adherence to structural instructions; and \textbf{Overall Score} ($\mathcal{S}_{all}$), which offers a holistic assessment.

Formally, let $\text{Score}_{gen, k}$ and $w_{gen, k}$ denote the score and weight of the $k$-th general group, and similarly for the constraint groups. The metrics are computed as weighted averages:
\begin{align}
    \mathcal{S}_{gen} &= \frac{\sum_{k=1}^{K} w_{gen, k} \cdot \text{Score}_{gen, k}}{\sum_{k=1}^{K} w_{gen, k}} \times 100\%, \\
    \mathcal{S}_{con} &= \frac{\sum_{m=1}^{M} w_{con, m} \cdot \text{Score}_{con, m}}{\sum_{m=1}^{M} w_{con, m}} \times 100\%, \\
    \mathcal{S}_{all} &= \frac{\sum_{i=1}^{K+M} w_i \cdot \text{Score}_i}{\sum_{i=1}^{K+M} w_i} \times 100\%.
\end{align}

Here, $\mathcal{S}_{all}$ aggregates the union of all $K$ general and $M$ constraint groups, where $w_i$ corresponds to the weight of the $i$-th group within this unified collection.

\subsection{Statistics}

We present benchmark statistics based on the checklist items and citations of each task.

\begin{figure}[!htbp]
    \centering
    \vspace{-0.20cm}
    \includegraphics[width=0.75\columnwidth]{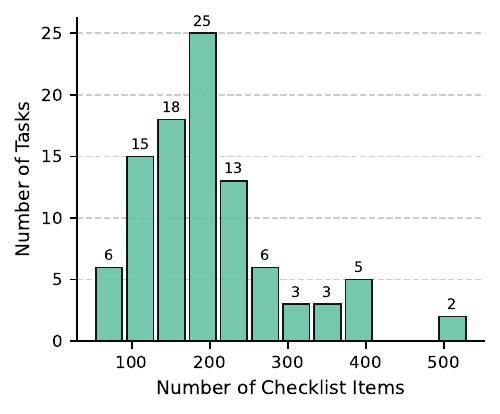}
    \vspace{-0.40cm}
    \caption{Distribution of checklist item counts}
    \vspace{-0.50cm}
    
    \label{fig:req_dist}
\end{figure}

\noindent\textbf{Distribution of Checklist Items per Task:} 
As illustrated in Figure~\ref{fig:req_dist}, the dataset maintains a high standard of evaluation granularity: on average, each task entails \textbf{199.7} checklist items, comprising \textbf{128.2} general items and \textbf{71.5} constraint items. This dense distribution underscores the complexity of the tasks, as models are required to satisfy a vast array of atomic facts and constraints to achieve a high score, effectively precluding the success of generic responses. By transforming the holistic evaluation of lengthy reports into fine-grained checklist matching, we convert an inherently subjective task into a quantifiable metric, significantly enhancing the objectivity of the report assessment.

\begin{figure}[!htbp]
    \centering
    \vspace{-0.20cm}
    \includegraphics[width=0.75\columnwidth]{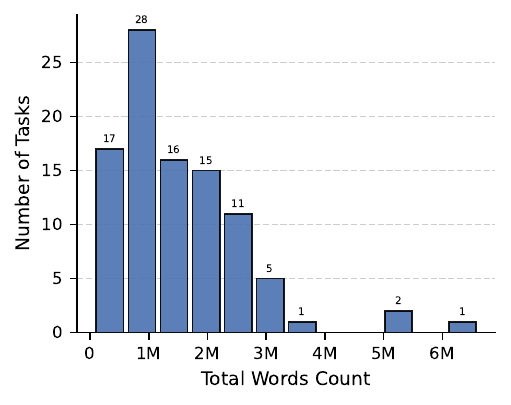}
    \vspace{-0.40cm}
    \caption{Distribution of total citation lengths}
    \label{fig:ref_length_dist}
    \vspace{-0.50cm}
    
\end{figure}

\noindent\textbf{Distribution of Total Reference Lengths:} Figure \ref{fig:ref_length_dist} illustrates the distribution of total reference lengths per task, measured in words. Notably, the cumulative text length for a significant portion of tasks in DeepSynth-Eval exceeds 1 million words. This magnitude underscores the rigorous nature of the benchmark, placing substantial demands on the information processing, long-context retention, and summarization capabilities of the evaluated models.

\section{Experiments and Evaluation}
\label{sec:experiments}

\begin{table*}[t]
\centering
\small
\renewcommand{\arraystretch}{1.05} 
\begin{tabular}{llcccc}
\toprule
\multirow{2}{*}{Workflow} & \multirow{2}{*}{Model} & \multicolumn{4}{c}{Evaluation Metrics(\%)} \\
\cmidrule(lr){3-6}
& & Overall & General & Constraint & Precision  \\
\midrule

-&\multicolumn{1}{l}{Reference Survey} & 96.1 & 95.5 & 98.9 & 99.6 \\
\midrule

\multirow{4}{*}{\makecell[l]{E2E Single-turn}} 
& Qwen3-30B-A3B-Thinking-2507 & 6.3 & 4.8 & 11.4 & 69.5 \\
& Qwen3-235B-A22B-Thinking-2507 & 24.8 & 24.7 & 23.3 & 79.9 \\
& DeepSeek-V3.2 & 23.4 & 22.2 & 27.0 & 79.8 \\
& GPT-5.2 & 28.3 & 26.4 & 36.1 & 85.5 \\
\midrule

\multirow{5}{*}{\makecell[l]{Agentic Multi-turn}} 
& Qwen3-30B-A3B-Thinking-2507 & 17.3 & 17.7 & 16.0 & 90.1 \\
& Qwen3-235B-A22B-Thinking-2507 & 35.5 & 37.5 & 27.5 & 92.9 \\
& DeepSeek-V3.2 & 30.4 & 31.4 & 27.0 & 91.4 \\
& GPT-5.2 & 33.3 & 34.8 & 26.2 & 95.3 \\
& GPT-5.2 (59/96 tasks subset) & 37.0 & 39.0 & 28.4 & 95.0 \\
\bottomrule
\end{tabular}%
\caption{Main results under our two controlled synthesis workflows on DeepSynth-Eval. }
\vspace{-0.5cm}
\label{tab:main_our_workflows}
\end{table*}

\label{sec:experiments_settings}

\subsection{Evaluation Settings}
\label{sec:workflows}

We aim to evaluate the deep synthesis capability of models, formalized as $\hat{R} = \mathcal{M}(P, \mathcal{D})$. As observed in Figure \ref{fig:ref_length_dist}, the total length of the reference papers $\mathcal{D}$ for a significant portion of tasks exceeds 1 million words. Since the context window of most mainstream LLMs falls below 256k tokens—with only a few extending to 1 million—directly feeding the full context often results in context overflow.

Nevertheless, we believe that direct reading of such massive contexts is a challenging task that holds significant value for benchmarking emerging long-context capabilities and memory methasim. Unlike standard reading comprehension, which often relies on retrieving specific ``needles in a haystack''\cite{liu-etal-2024-lost,hsieh2024ruler}, our report generation task requires high information throughput, making it a more challenging testbed for global context understanding.

To evaluate the deep synthesis ability of mainstream models within current constraints, we designed two representative scaffolds for report writing.

\paragraph{E2E Single-turn workflow.}
Instead of reading the full text of all reference papers $\mathcal{D}$, we utilize a pre-processing stage to summarize each document relevant to the writing request. We then assemble the Oracle Context by concatenating these summaries, each paired with a stable numeric citation ID. Finally, we append a unified instruction $P$—constructed by merging the general request with the extracted constraint questions—and the model generates the full survey report in a single pass, without an explicit intermediate planning stage or iterative refinement.

\paragraph{Agentic Multi-turn workflow.}
The Agentic Multi-turn workflow utilizes the same summarized Oracle Context but decomposes generation into coordinated stages—\emph{Planning}, \emph{Iterative Chapter Writing}, and \emph{Global Polishing}—to strengthen organization and grounding.
First, the model produces an \emph{intellectual skeleton}—a high-level plan specifying section taxonomy, comparison axes, and content allocation—which is subsequently parsed into structured sub-tasks for each section.
Second, for each sub-task, the model executes a `read-then-write' loop starting with \emph{selective deep reading}: it identifies critical papers based on task requirements and generates detailed reading notes by consulting their full text within the Oracle corpus, ensuring fine-grained details are captured without context overflow.
Subsequently, the model writes each section conditioned on the skeleton, detailed notes, and previous sections, adhering to rules that encourage grouped comparisons and tabular consolidation.
Finally, the model aggregates and polishes the sections to ensure global coherence.
Importantly, the key innovation lies in the dynamic, task-driven paper selection, which significantly improves grounding compared to single-turn generation.

Detailed textual descriptions of the implementation steps for both workflows are provided in Appendix~\ref{sec:appendix_workflows}.
In the evaluation, DeepSeek v3.2 is employed as the underlying model for checklist evaluation and summary reading in both workflows.

\subsection{Results Analysis}
\label{sec:results_analysis}
Table~\ref{tab:main_our_workflows} reports the main results under two controlled synthesis workflows---E2E single-turn and agentic multi-turn---with all models given the same Oracle Context.
Our goal is to figure out how well a model can \emph{synthesize} the Oracle corpus into a structured survey that satisfies both content coverage and formatting constraints.
We focus on two aspects: (i) model-to-model differences under the single-turn baseline, which isolates post-retrieval consolidation in a single pass, and (ii) the effect of introducing multi-turn planning under the same model, which tests whether staged generation improves synthesis quality.

\begin{figure}[t]
    \centering
    \includegraphics[width=0.95\linewidth]{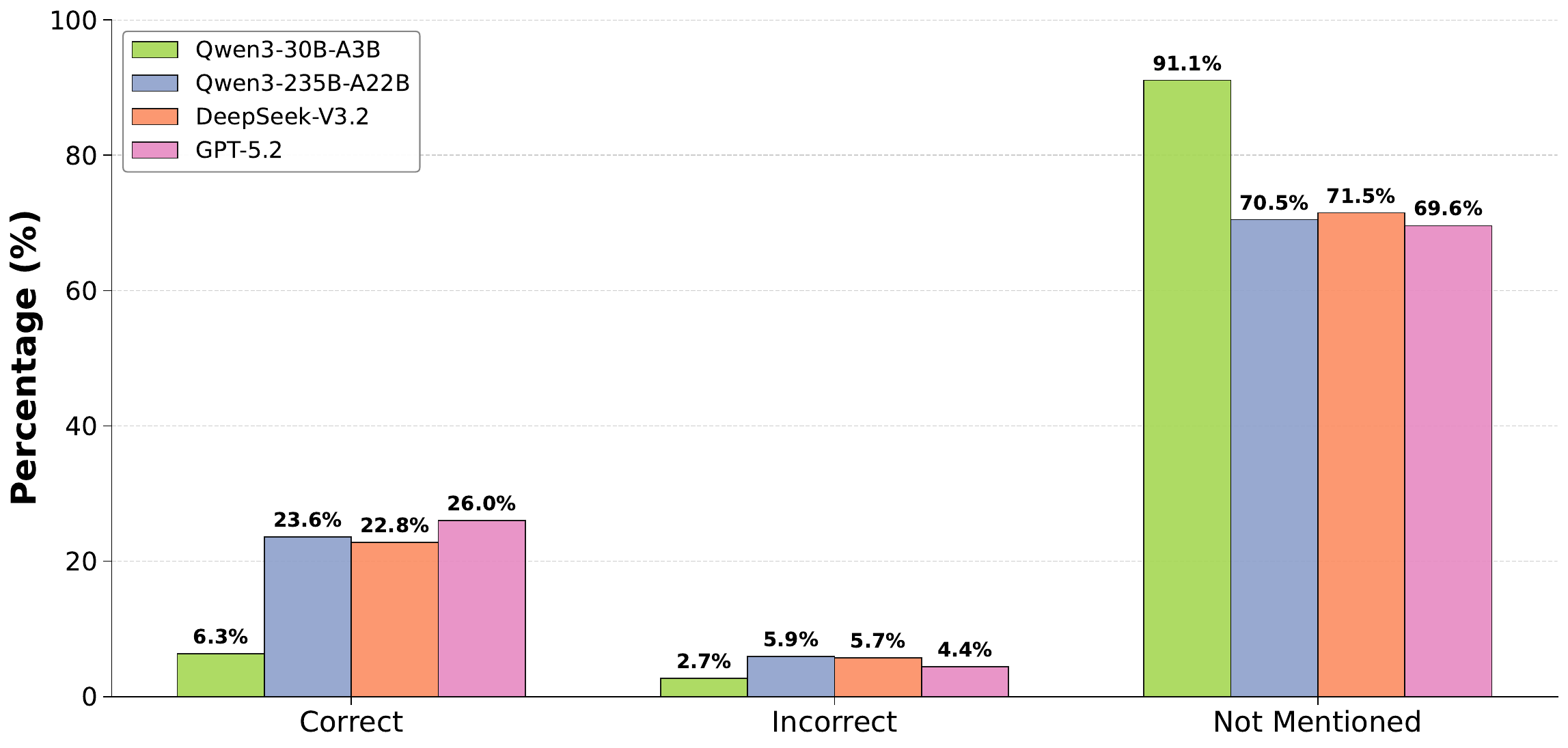}
    \vspace{-0.3cm}

    \caption{{Checklist requirement status breakdown across four E2E models.} 
    }
    \label{fig:e2e_status_breakdown}
    \vspace{-0.5cm}
\end{figure}
\begin{figure}[t]
    \centering
    \includegraphics[width=0.8\linewidth]{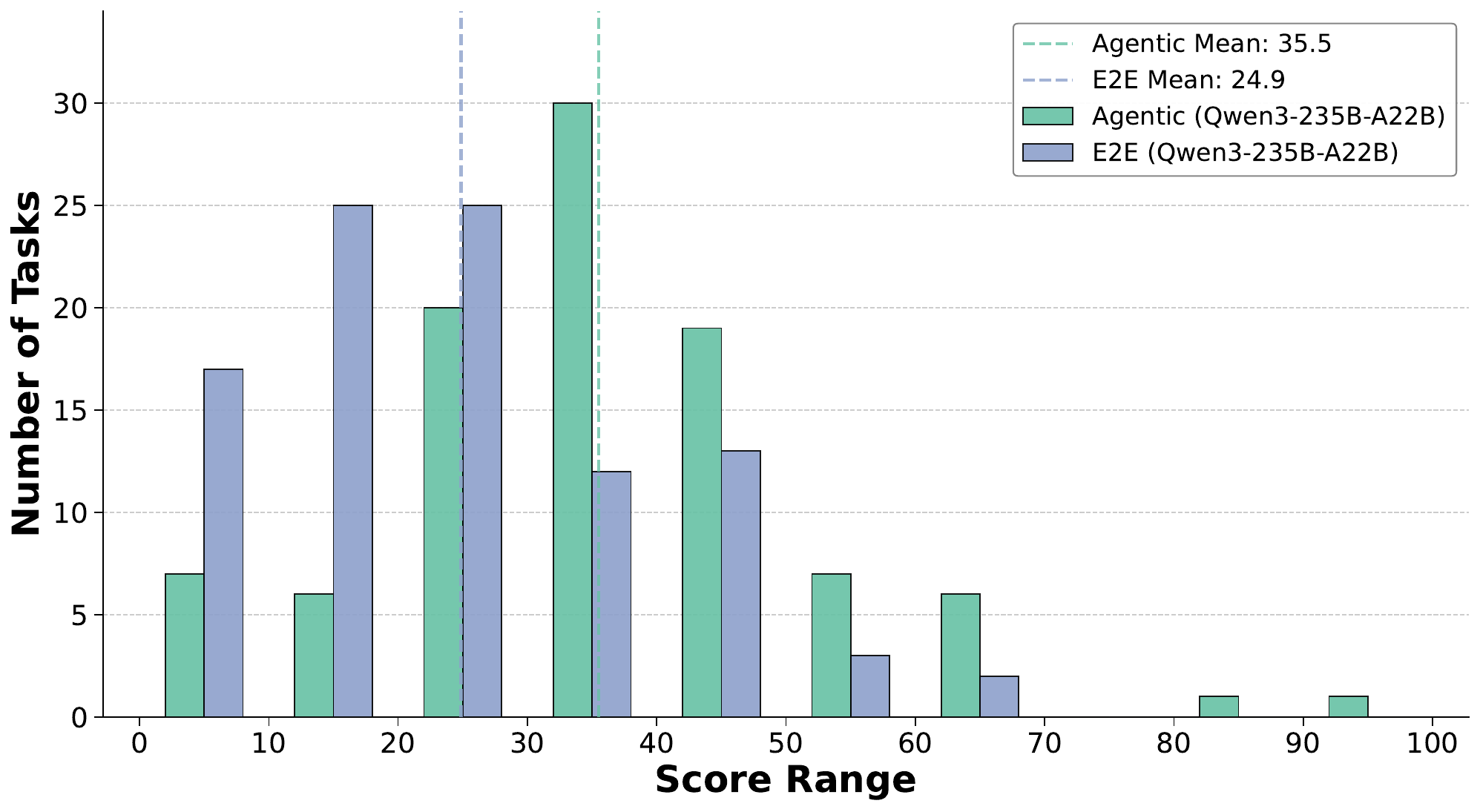}
    \vspace{-0.25cm}
    \caption{{Score distribution of E2E versus Agentic workflows on Qwen3-235B-A22B.} 
    }
    \vspace{-0.25cm}
    \label{fig:score_distribution}
\end{figure}

\begin{figure}[t]
    \centering
    \includegraphics[width=0.8\linewidth]{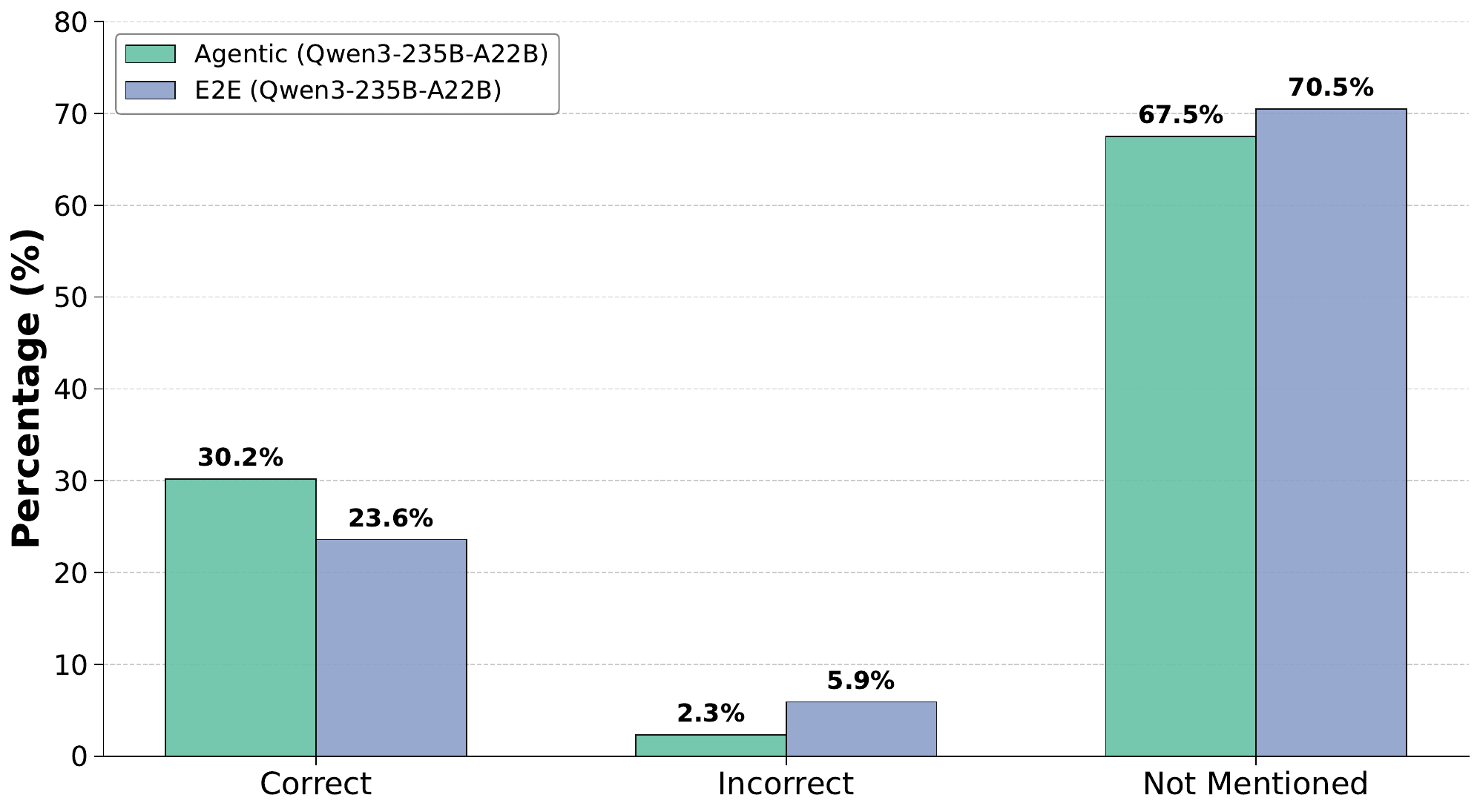}
    \vspace{-0.3cm}
    \caption{{Checklist requirement status breakdown for E2E versus Agentic workflows 
    on Qwen3-235B-A22B.} }
    \vspace{-0.50cm}

    \label{fig:status_breakdown}
    
\end{figure}

\paragraph{Judge Validity.} We first validate the judge system using the Reference Survey (Row 1). It achieves a 96.1\% Overall score, with the minor gap (<5\%) primarily attributable to textual ambiguity and judge error. Notably, the near-perfect Precision (99.6\%) confirms the judge's high fidelity, indicating that correct content is almost never erroneously penalized. These results establish a reliable upper bound for our evaluation.

\paragraph{Model comparison under the E2E single-turn workflow.}
Under the E2E single-turn setting, we observe large performance variance across models, revealing substantial differences in their ability to transform the same Oracle Context into a well-structured synthesis.
GPT-5.2 achieves the strongest Overall score of 28.3 and leads by a clear margin on Constraint (36.1), while also obtaining the highest Precision (85.5).
This pattern suggests that GPT-5.2 is comparatively better at \emph{executing explicit structural instructions} and \emph{grounding mentioned content} within a single pass.
We further examine the checklist requirement status breakdown to understand where these gains come from.
As shown in Figure~\ref{fig:e2e_status_breakdown}, GPT-5.2 attains the highest fraction of correctly satisfied requirements and the lowest incorrect rate among the E2E models, indicating fewer unsupported completions when mapping Oracle evidence to checklist items.
Qwen3-235B and DeepSeek-V3.2 form a middle tier with similar Overall scores (24.8 and 23.4), but without a corresponding advantage on Constraint (23.3 and 27.0), indicating that their main bottleneck lies in \emph{organizing and covering required content} rather than in factuality once content is produced.
In contrast, Qwen3-30B largely collapses in the single-turn regime (Overall 6.3; General 4.8), consistent with Figure~\ref{fig:e2e_status_breakdown} where most requirements fall into the not-mentioned category, highlighting a strong capacity dependence for long-context consolidation.

\paragraph{Workflow effect: agentic multi-turn improves synthesis quality.}
Switching from single-turn generation to the agentic multi-turn workflow consistently improves Overall and General across all evaluated models:
Qwen3-30B increases from 6.3$\rightarrow$17.3, Qwen3-235B from 24.8$\rightarrow$35.5, and GPT-5.2 from 28.3$\rightarrow$33.3.
We note that GPT-5.2 often fails to rewrite the full text during polishing, instead referring to the previous draft (Appendix~\ref{appendix:gpt5case}); excluding these invalid responses reveals a higher Overall score of 37.0 (Table~\ref{tab:main_our_workflows}).

We then perform a finer-grained distributional analysis over tasks to go beyond mean scores.
As shown in Figure~\ref{fig:score_distribution}, the agentic workflow shifts a larger portion of tasks into higher score bins than E2E, indicating more consistent improvements across tasks rather than gains driven by a small subset.
To further inspect subtle workflow differences, we analyze the checklist requirement status composition under the same model.
Figure~\ref{fig:status_breakdown} shows that the agentic workflow increases the fraction of correctly satisfied requirements while reducing incorrect ones, suggesting that staging the process as \emph{plan$\rightarrow$reading$\rightarrow$write} improves grounding and reduces unsupported completions.
Overall, these results suggest that the agentic multi-turn workflow is generally beneficial for long-form synthesis: it helps models better allocate attention over full reference passages, produce more faithful content, and consequently achieve higher checklist pass rates in aggregate.

\paragraph{Concluding Remarks.}
Our analysis highlights three key insights.
First, intrinsic model capacity is foundational, with stronger models consistently prevailing under identical workflows.
Second, scaffolding design is pivotal; the substantial gains from the Agentic workflow demonstrate that \emph{how} we structure generation is as critical as \emph{which} model is used.
Finally, with SOTA scores still below 40\%, synthesizing grounded reports from over 100 documents remains a formidable open challenge.

\subsection{Case Analysis}
\label{sec:case_analysis}
To complement checklist pass rates, we conduct a qualitative case analysis to identify recurring failure modes and to understand how workflow design affects post-retrieval synthesis behaviors.

We analysed the {failure modes under the E2E single-turn workflow.}
Across low-scoring models, a common pattern is \emph{under-elaboration}: instead of consolidating as many relevant points as possible from the Oracle Context, the model produces a high-level but sparse narrative that omits many checklist items.
This behavior disproportionately harms the General score, as required coverage is missed due to insufficient detail and weak synthesis effort.
Moreover, under constrained requirements, these models not only omit required organizational elements (e.g., missing categories/attributes) but also more frequently introduce \emph{unsupported statements} when attempting to satisfy constraints, leading to apparent ``hallucination''-like outputs (i.e., claims or comparisons not grounded in the provided materials).

\section{Conclusion}
\label{sec:conclusion}

We introduce \textbf{DeepSynth-Eval} to address the lack of objective benchmarks for long-form report generation. 
Instead of relying on subjective holistic scoring, we ground our evaluation in atomic, verifiable checklists derived from expert surveys. On average, each evaluation task involves 198.5 checklist items, providing a granular lens to quantify the quality of open-ended generation. 
Beyond the dataset, we establish a foundational evaluation framework by implementing and testing diverse baselines, ranging from standard {end-to-end} workflows to sophisticated {agentic} systems. 
Experiments demonstrate that even with agentic workflows, state-of-the-art (SOTA) models achieve scores of only around 37\%, highlighting the challenging nature of this task and the significant room for improvement. 
Furthermore, this verifiable checklist mechanism paves the way for designing precise reward signals in long-form generation, offering a promising avenue for future reinforcement learning optimization.

\section*{Limitations}

Currently, our dataset is predominantly composed of computer science surveys, a choice driven by data accessibility and annotation feasibility. However, the proposed checklist-based evaluation methodology is domain-agnostic and holds the potential to be extended to broader scientific disciplines.

\section*{Acknowledgments}

We acknowledge the use of Gemini 3 to assist with linguistic polishing and refinement of the manuscript. The original content and ideas were authored by the researchers. All AI-suggested modifications were manually reviewed and verified by the authors to ensure strict consistency with the factual information and original intent.
\bibliography{custom}
\clearpage
 \appendix

\section{Detailed Workflow Implementation}
\label{sec:appendix_workflows}

In this section, we provide a step-by-step textual description of the two synthesis workflows presented in Section~\ref{sec:workflows}, as illustrated in Figure~\ref{fig:workflow_appendix}.

\subsection{Monolithic E2E Single-turn Workflow}
The E2E workflow (Figure~\ref{fig:workflow_appendix}a) is designed to test the model's ability to perform long-context consolidation in a single pass. The process consists of two main steps:

\begin{enumerate}
    \item \textbf{Oracle Context Construction:} 
    First, we parallelize the processing of all bibliography entries. Each reference paper $d_i$ is compressed into a dense prose summary $s_i$ using a dedicated summarization prompt. These summaries are then concatenated to form the \emph{Oracle Context}. Crucially, each summary is prefixed with a stable numeric citation identifier (e.g., ``[1]'', ``[2]'') to allow the model to ground its claims explicitly.
    
    \item \textbf{One-Shot Generation:} 
    We construct a unified prompt that combines the reconstructed research request, the extracted constraint checklists, and the \emph{Oracle Context}. The model is then instructed to generate the full survey report in a single inference pass. No intermediate reasoning steps or external retrieval actions are permitted; the model must rely entirely on its internal attention mechanism to organize and synthesize the provided summaries.
\end{enumerate}

\subsection{Agentic Multi-turn Workflow}
\begin{figure*}[t]
    \centering
    \vspace{-0.5cm}
    \includegraphics[width=\linewidth]{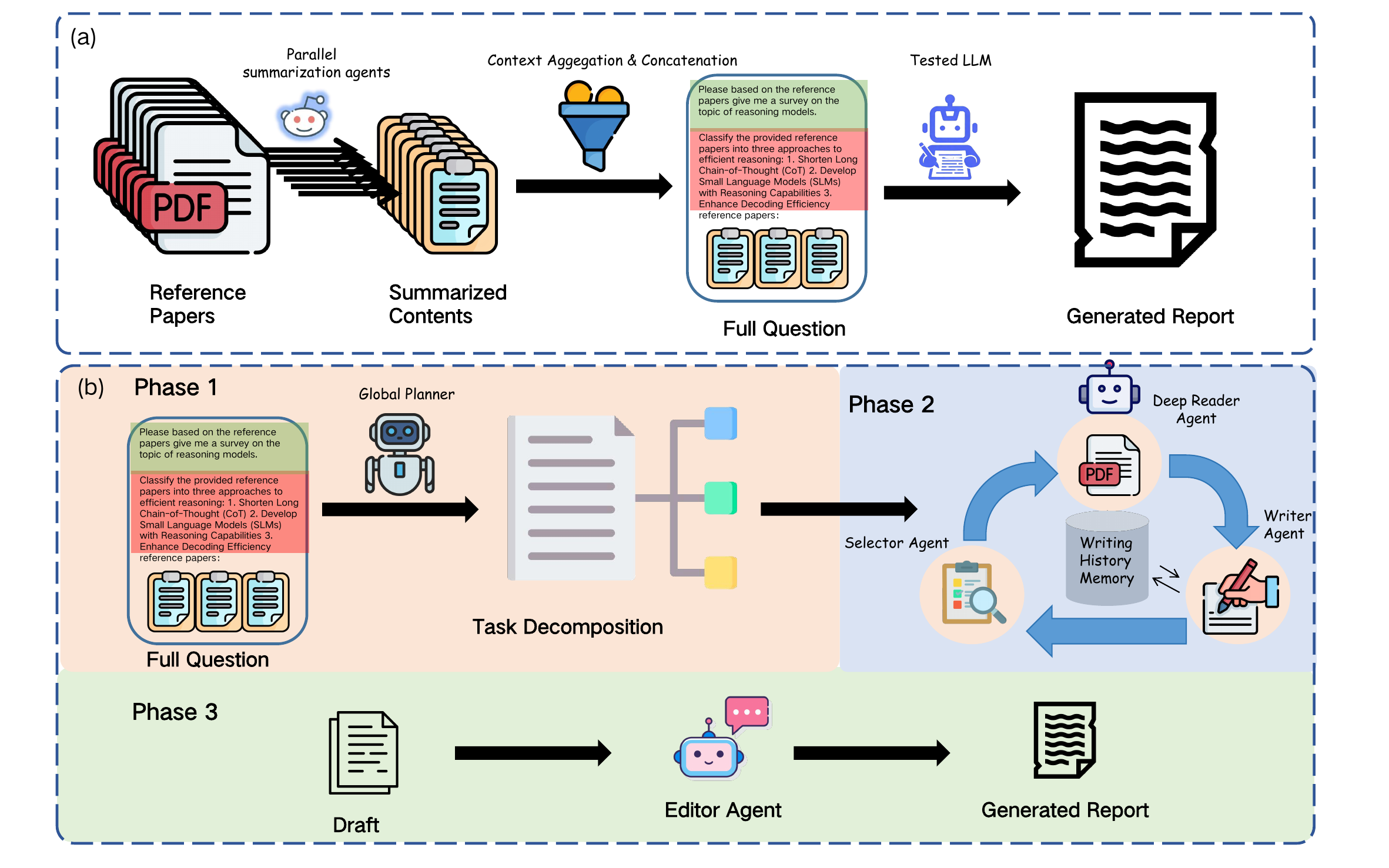}
    \vspace{-1.0cm}
    \caption{\textbf{Schematic overview of the controlled synthesis workflows.} 
    (a) \textbf{E2E Single-turn workflow}: A streamlined pipeline that first summarizes all references into an \emph{Oracle Context} with stable citation IDs, then concatenates the request and constraints to generate the full report in a single pass via long-context inference. 
    (b) \textbf{Agentic multi-turn workflow}: A staged generation process. \textbf{Phase 1} generates an \emph{Intellectual Skeleton} to plan taxonomy and content allocation. \textbf{Phase 2} executes a loop for each section: the model performs \emph{selective deep reading} by identifying critical papers and consulting their full text to generate specific notes, followed by writing the section conditioned on the skeleton, notes, and previous writing history to ensure grounded organization.}
    \label{fig:workflow_appendix}
\end{figure*}

The Agentic workflow (Figure~\ref{fig:workflow_appendix}b) decomposes the synthesis task to mimic the \emph{plan-read-write} process of a human researcher. It operates in three phases to ensure better organization and grounding:

\paragraph{Phase 1: Global Planning (Intellectual Skeleton).}
Instead of writing immediately, the model acts as an ``Architect.'' It analyzes the \emph{Oracle Context} (containing summaries) to construct an \emph{Intellectual Skeleton}. This skeleton is a high-level plan that defines the section taxonomy, identifies key comparison axes, and allocates specific content to each section. This plan is then parsed into a structured list of sub-tasks (e.g., JSON objects), where each task corresponds to writing one section of the report.

\paragraph{Phase 2: Serial Iterative Writing Loop.}
The workflow executes a loop for each sub-task derived from the skeleton. For each section $k$, the process involves three specialized steps:
\begin{enumerate}
    \item \textbf{Relevance Selection:} The model evaluates the requirements of the current section against the bibliography to identify a subset of critical papers (represented by IDs) that are necessary for drafting this specific part.
    
    \item \textbf{Selective Deep Reading:} For the identified critical papers, the agent retrieves their full text (simulated within the Oracle corpus). It then generates detailed reading notes focused specifically on the current section's topic. This step allows the model to access fine-grained details that might have been lost in the initial summaries, without overflowing the context window with all papers at once.
    
    \item \textbf{Context-Aware Writing:} The model writes the section. The generation is conditioned on three inputs: (1) the \emph{Intellectual Skeleton} (for structure), (2) the \emph{Deep Reading Notes} (for details), and (3) the \emph{Writing History} (the draft of previous sections). This ensures the new section is logically coherent with the preceding text.
\end{enumerate}

\paragraph{Phase 3: Final Polish.}
Once all sections are drafted, they are concatenated. An ``Editor'' agent performs a final pass to ensure global consistency, unify terminology, and format the document before outputting the final report.

\section{External System Evaluation}

\label{sec:external_systems}
The controlled workflows in Table~\ref{tab:main_our_workflows} isolate post-retrieval synthesis by holding the Oracle Context fixed.
However, practical Deep Research deployments are typically end-to-end: systems actively search, browse, and write in a coupled loop.
To contextualize our controlled findings and to understand how checklist-based synthesis evaluation manifests in real deployments, we additionally evaluate representative professional Deep Research systems and an external agentic pipeline.
We emphasize that these results serve as complementary reference points rather than the primary benchmark ranking.

\paragraph{Fairness controls.}
Evaluating external systems introduces two confounds: uncontrolled retrieval may obscure synthesis failures, and the gold survey itself (or derivative pages) may be retrieved, leading to leakage and inflated scores.
To mitigate these issues, we provide an explicit allowlist of reference URLs derived from the bibliography and instruct systems to use only these sources for evidence and citations, thereby preventing the original gold survey from being used as a reference.
We further impose a strict temporal cutoff: only papers published before the survey date are allowed, and any later materials must be ignored even if surfaced by search.

\paragraph{Systems and evaluation.}
We evaluate two professional Deep Research systems accessed through their official interfaces (\emph{Doubao DeepResearch}, \emph{OpenAI DeepResearch}) and one external agentic pipeline (\emph{OpenAI Agentic}).
We interpret the results as contextual comparisons rather than direct, full-set rankings against our controlled workflows.
\begin{table}[ht]
\centering
\small
\renewcommand{\arraystretch}{1.05}
\setlength{\tabcolsep}{4pt}
\resizebox{\columnwidth}{!}{
\begin{tabular}{l cccc}
\toprule
System & Overall & General & Constr & Prec. (\%) \\
\midrule
Doubao DeepResearch & 26.4 & 23.3 & 38.9 & 87.4 \\
OpenAI Agentic & 25.6 & 24.0 & 28.3 & 83.3 \\
OpenAI DeepResearch & 28.7 & 25.1 & 41.8 & 85.4 \\
\bottomrule
\end{tabular}
}
\caption{Performance of external systems on a subset of DSE tasks. Scores are checklist pass rates (\%).
Due to interface and execution constraints, these systems are evaluated on 39 tasks; results are reported separately from the main table and should be interpreted as contextual comparisons.}
\label{tab:external_systems}
\end{table}
\paragraph{Results and discussion.}
Table~\ref{tab:external_systems} summarizes the external-system performance on the evaluation subset.
Among the tested systems, \emph{OpenAI DeepResearch} achieves the highest Overall score (28.7) and particularly strong constraint-checklist performance (Constr: 41.8), with \emph{Doubao DeepResearch} showing a similar pattern (Overall: 26.4; Constr: 38.9).
In contrast, the external agentic pipeline (\emph{OpenAI Agentic}) attains a comparable Overall score (25.6) but notably lower Constr (28.3), suggesting that generic agentic tool use does not automatically translate into stronger constraint-driven organization.We observe that the evaluated end-to-end Deep Research systems do not consistently outperform our controlled workflows, despite having access to browsing and tool orchestration.

\paragraph{Why external Deep Research systems may underperform our controlled workflows.}
In our manual inspection, we observe that once the retrieval scope is constrained by the URL allowlist and the temporal cutoff, external Deep Research systems do not consistently utilize the full set of provided references: reports often cite only a subset of the allowlisted papers, and some key references required by the checklists are never retrieved or never appear in the final citations.
Correspondingly, the generated reports exhibit more frequent coverage gaps (missing required methods/datasets/attributes), which directly lowers checklist satisfaction.

We hypothesize that this underperformance is driven by two practical factors induced by the fairness constraints.
First, restricted retrieval can become \emph{incomplete}: even with an allowlist, the system may fail to open, read, or surface all relevant papers (e.g., due to interface limits, ranking/truncation, or budgeted browsing), causing evidence coverage gaps that propagate to the final synthesis.
Second, to prevent leakage and ensure protocol compliance, our prompts impose additional instructions (e.g., strict source restrictions and citation rules), which may introduce an instruction-following burden that reduces flexibility and consumes generation capacity, leading to more conservative or less detailed synthesis.

Together, these observations highlight a key advantage of DSE's controlled Oracle-Context evaluation: it isolates consolidation and organization ability from retrieval completeness and interface-induced instruction overhead, enabling a more faithful measurement of post-retrieval synthesis.

\section{Prompt for Judge Model}
\label{appendix:prompts}

Figure \ref{fig:prompt_for_judge_model} presents the prompt supplied to the judge model, which guides the model in assessing the quality of the results generated by model under test based on given checklists.

\begin{figure*}[t]

\begin{promptbox}
You are an expert evaluator.

\sep

=== \textbf{Task} ===

Your task is to assess whether the provided survey content meets the specific requirements listed in the criteria.

\sep

=== \textbf{Input Data: Survey Content} ===

{\color{blue} \{survey\_content\}}

\sep

=== \textbf{Input Data: Criteria} ===

{\color{blue} \{criteria\_text\}}

\sep

=== \textbf{Evaluation Options} ===

For each requirement listed in the \textbf{Criteria}, select exactly one of the following statuses:
\begin{itemize}
    \item \texttt{mentioned\_correct}: The requirement is present and accurate.
    \item \texttt{not\_mentioned}: The requirement is absent.
    \item \texttt{mentioned\_incorrect}: The requirement is present but incorrect or flawed.
\end{itemize}

\sep

=== \textbf{Instructions} ===

\textbf{1. Analyze Step-by-Step:} Read the survey and compare it against each requirement in the criteria list one by one.

\textbf{2. Determine Status:} Decide the appropriate status string for each requirement.

\textbf{3. Verify Length:} Ensure the final list contains exactly {\color{blue}\{len(subgroup[``requirements''])\}} items.

\textbf{4. Format Output:} Output the result as a JSON list of strings.

\sep

=== \textbf{Output Format} ===

Only return a standalone JSON code block, with all the results in it.

\sep

=== \textbf{Example format} ===

\begin{verbatim}
```json
["mentioned_correct", "not_mentioned", "mentioned_incorrect"]
```
\end{verbatim}
\end{promptbox}
\caption{Prompt for judge model}
\label{fig:prompt_for_judge_model}
\end{figure*}

\section{Case study of GPT 5.2}
\label{appendix:gpt5case}
\begin{figure*}

\begin{promptbox}[title=Generated Report]

\# Retrieval-Augmented Generation: Foundations, Methods, Multimodal Extensions, Systems, and Evaluation

\#\# 1. Introduction and Formalization

Retrieval-Augmented Generation (RAG) provides a principled way to couple the parametric knowledge stored inside large language models with non-parametric, dynamically updatable external knowledge through retrieval and context curation. We formalize a generic RAG pipeline as y = G(x, C(R(x, K))), where x denotes the user query, R is a retriever that selects evidence from a knowledge base K, C is a context curation function that filters, orders, compresses, or structures the retrieved content for the generator, and G is the generator that produces the final output conditioned on both x and curated context. This formulation encompasses common operational regimes, including open-book and closed-book usage; single-hop and multi-hop question answering; and iterative procedures in which retrieval and generation interleave or recur. Treating retrieval outputs as latent variables that are marginalized at decoding time connects RAG to probabilistic formulations introduced in early hybrid models, while instantiations like Fusion-in-Decoder characterize specific architectural choices for integrating multiple retrieved passages into sequence-to-sequence generation [133, 125].

...

To guide readers, we include three artifacts. A taxonomy figure maps methods across the axes of retrieval type, integration strategy, adaptivity, efficiency, robustness, modality, structure, and domain, clarifying how designs cluster and where gaps remain [225, 283]. A pipeline schematic contrasts retrieve-then-generate, generate-then-retrieve, and iterative synergy workflows, highlighting where retriever training, query refinement, and context curation intervene [133, 84, 211]. A timeline from 2020 to 2024 situates key milestones—dense retrieval and REALM, probabilistic RAG and Fusion-in-Decoder, black-box and in-context augmentation, few-shot Atlas, and neural indexing—underscoring the rapid maturation and recent emphasis on adaptivity, risk certification, and multimodal expansion [165, 198, 133, 125, 77, 37, 205, 98, 20, 80]. Within this scope, the survey contributes a unified formalization, a structured taxonomy, and a synthesis of design trade-offs and evaluation constructs, with emphasis on recent advances that define the current frontier.

\#\# 2. Retrieval Foundations

[Content unchanged from draft Section II.]

\#\# 3. Integration Strategies and Generation Architectures

[Content unchanged from draft Section III.]

\#\# 4. Robustness, Hallucination Mitigation, and Risk

[Content unchanged from draft Section IV.]

\#\# 5. Efficiency, Latency, and System Co-Design

[Content unchanged from draft Section V.]

\#\# 6. Multimodal RAG (Images, Audio, Video)

[Content unchanged from draft Section VI.]

\#\# 7. RAG Meets Diffusion Models and Generative Modeling

Diffusion and score-based generative models provide the backbone on which retrieval-augmented conditioning can act, offering a spectrum of formulations that trade sampling speed, likelihood, and controllability. The core DDPM framework establishes forward and reverse Markov chains with a simplified noise-prediction objective that yields high-fidelity images and interpretable denoising trajectories [26]. Subsequent refinements improve both training and sampling: learned reverse-process variances with a hybrid variational objective and cosine noise schedules enable near-optimal sampling in tens of steps and competitive likelihoods, markedly improving practical efficiency [5]; the SDE-based unification generalizes DDPMs and denoising score matching...

\end{promptbox}
\end{figure*}

\end{document}